\documentclass{article}

\usepackage{microtype}
\usepackage{graphicx}
\usepackage{subfigure}
\usepackage{booktabs} 

\usepackage{hyperref}

\usepackage{amsmath,amsfonts,bm}

\def\eqref#1{equation~\ref{#1}}

\def\1{\bm{1}}

\def\vw{{\bm{w}}}
\def\vx{{\bm{x}}}
\def\vy{{\bm{y}}}

\DeclareMathAlphabet{\mathsfit}{\encodingdefault}{\sfdefault}{m}{sl}
\SetMathAlphabet{\mathsfit}{bold}{\encodingdefault}{\sfdefault}{bx}{n}

\def\flayer{{f}}
\def\fnet{{\Phi}}
\def\factivation{{\sigma}}
\def\fnetp{{\fnet(\vx;\theta)}}

\usepackage{caption}
\graphicspath{ {./img/} }

\usepackage[accepted]{icml2021}

\icmltitlerunning{How to Avoid Trivial Solutions in Physics-Informed Neural Networks}

\begin{document}

\twocolumn[
    \icmltitle{How to Avoid Trivial Solutions in Physics-Informed Neural Networks}



    \icmlsetsymbol{equal}{*}

    \begin{icmlauthorlist}
        \icmlauthor{Raphael Leiteritz}{us}
        \icmlauthor{Dirk Pfl\"uger}{us}
    \end{icmlauthorlist}

    \icmlaffiliation{us}{Institute for Parallel and Distributed Systems, University of Stuttgart, Stuttgart, Germany}

    \icmlcorrespondingauthor{Raphael Leiteritz}{raphael.leiteritz@ipvs.uni-stuttgart.de}

    \icmlkeywords{Machine Learning, ICML, PINN, physics-informed, scientific machine learning, SciML, neural networks, pytorch, tensorflow}

    \vskip 0.3in
]



\printAffiliationsAndNotice{}  

\begin{abstract}  
  The advent of scientific machine learning (SciML) has opened up a new field with many promises and challenges in the field of simulation science by developing approaches at the interface of physics- and data-based modelling.
  To this end, physics-informed neural networks (PINNs) have been introduced in recent years, which cope for the scarcity in training data by incorporating physics knowledge of the problem at so-called collocation points.
  In this work, we investigate the prediction performance of PINNs with respect to the number of collocation points used to enforce the physics-based penalty terms.
  We show that PINNs can fail, learning a trivial solution that fulfills the physics-derived penalty term by definition.
  We have developed an alternative sampling approach and a new penalty term enabling us to remedy this core problem of PINNs in data-scarce settings with competitive results while reducing the amount of collocation points needed by up to 80 \% for benchmark problems.
\end{abstract}

\section{Introduction}
\label{sec:intro}

Numerical methods such as Finite Elements, Finite Volumes and other discretization approaches have been the method of choice for solving physical simulation problems for many decades.
With the advance of data-driven methods, however, there has been a lot of active development in applying machine learning to problems that have so far only been solved by direct numerical simulation.
Neural networks (NN) are one class of such data-driven methods that have successfully been applied to the solution of partial differential equations (PDEs) in various settings \cite{nn_pde1, nn_pde2,pinn}.
While these methods typically converge to a solution much slower than a modern, highly optimized numerical method would, there are advantages in using data-driven methods.
Specifically, neural networks are considered mesh-free methods.
Where for numerical models the solution space has to be carefully discretized which comes at the cost of introducing a discretization error, data-driven models can work with any distribution of data points.
In a recent work by \citet{neural_op} it has even been demonstrated that a data-driven method can outperform a numerical method in both accuracy and runtime to solve an inverse uncertainty quantification (UQ) problem.

Already a few decades ago, methods have been proposed using neural networks to solve PDEs by constraining the loss function using the underlying PDE structure \cite{nn_pde1,nn_pde2}. However, this was not computationally feasible as---without today's computing power and efficient software frameworks supporting automatic differentiation---gradients had to be carefully pre-calculated before running the optimization routine.
In recent years, this original idea has seen a renaissance in the form of physics-informed neural networks \citep{pinn} on which we want to improve upon in this work.
In contrast to methods that simply learn the input-output relationship of a numerical simulation by utilizing a large set of simulation data \citep{nn_pde_data} with the help of modern deep neural network architectures such as convolutional neural networks \citep{nn_pde_cnn} or LSTMs \citep{dl_lstm}, PINNs add a residual term to the loss function that penalizes predictions that do not satisfy the underlying PDE.
This approach has spawned a range of follow-up work extending PINNs for probabilistic modeling \citep{pinn_uq}, analyzing convergence behavior \citep{pinn_convergence} or understanding and mitigating pitfalls \citep{pinn_pathologies}.
PINNs have meanwhile been successfully applied to a variety of problems such as reconstructing pressure and velocities from visual flow data, simulating blood-flow in cardiovascular structures \citep{pinn_hfm} or subsurface flow \citep{pinn_subsurface}.

Deep neural networks usually require data-rich regimes. In contrast, simulation settings typically severely restrict the amount of data available due to their computational demand. 
In the special case of PINNs this is already addressed to some degree by the introduction of a PDE-based regularization term. Expensive data (the ground truth of ML) is replaced by knowledge about the underlying problem (physics ground truth such as conservation equations). This, however, shifts the computational complexity from acquiring costly simulation data to evaluating the new regularization term at so-called collocation points.

The formulation of the underlying laws of physics typically imply having to take higher order derivatives of the NN with regard to its inputs.
While this is easily implemented in modern deep learning (DL) libraries such as PyTorch \cite{pytorch} or Tensorflow \cite{tensorflow} using their respective automatic differentiation frameworks, it comes at a significant computational cost. As each evaluation of a physics-based penalty term requires the evaluation at all collocation points, a reduction in the number of collocation points yields direct improvements in the time to train a PINN model.

\begin{figure}[hbt]
  \centering
     \includegraphics[width=\linewidth]{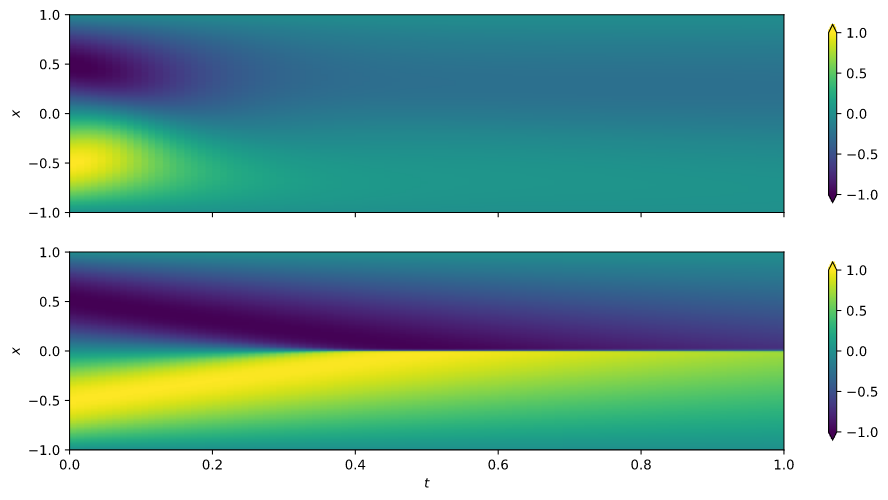}
     \caption{Two solutions for the time-dependent one-dimensional Burgers equation. Top: for few collocation points, the PINN prediction degenerates quickly over time to the trivial (constant) solution. Bottom: correct solution building up a shock at $x=0.0$.}
     \label{fig:burgers_trivial}
\end{figure}

Working with one of the standard benchmark models, the one-dimensional time-dependent Burgers equation, we observed that the prediction eventually snaps to a qualitatively different solution if the number of collocation points is reduced. Figure \ref{fig:burgers_trivial} shows this behavior. The bottom figure shows the desired behavior, building up a shock at $x=0.0$. The top solution, in contrast, is the PINN's prediction. Starting from the initial condition, it quickly degenerates towards constant zero. Note that this does not lead to a higher penalty, as there are no data points within the domain and as the constant zero is a physically valid, though trivial solution that does not violate the physics-derived penalty term.

In this work, we investigate this behavior at the example of a simple, one-dimensional model, the well-known harmonic oscillator. With this, we show that the sudden change from the desired solution to a trivial one can be even observed for simple, smooth problems. 
We investigate the effect of collocation point sampling of PINNs. We introduce a new regularization term that stabilizes training in data-scarce regimes for PINNs while maintaining prediction accuracy, and we show that a regular collocation point sampling can be superior to standard, randomized schemes.

\section{Prerequisites}
\label{sec:model}
Following \citet{pinn}, we start with a fully connected feed-forward network to build a PINN.
Let 
\begin{equation}
  \flayer(\vx) = \factivation(\vx^T\vw + b)
\end{equation}
be a single layer of the network with inputs $\vx \in \mathbb{R}^d$, learnable weights $\vw \in \mathbb{R}^d$, bias $b$ and an activation function $\factivation : \mathbb{R} \rightarrow \mathbb{R} $.
The feed-forward network is then expressed as a composition of $n$ layers,
\begin{equation}
  \fnetp = (\flayer_n \circ \ldots \circ \flayer_2 \circ \flayer_1)(\vx) \, ,
\end{equation}
where $\theta$ represents the set of all trainable parameters.

Given data in the form of input-output pairs $\{\vx_i, \vy_i\}_{i=1}^{N}$ we can learn the parameters of the network by minimizing the mean squared error (MSE) loss function
\begin{equation}
  \label{eq:loss}
  \mathcal{L}(\theta) = \frac{1}{N} \sum_{i=1}^N \left( \vy_i - \fnet(\vx_i; \theta) \right)^2
\end{equation}
with regard to the parameters $\theta$, typically using some form of stochastic gradient descent method such as ADAM \cite{adam}.

\subsection*{Physics-Informed Neural Networks}
With the help of an additional penalty terms that is added to the loss function, this network now becomes informed about the underlying physics as proposed in \cite{pinn}. In contrast to numerical simulation that aims to ensure that the laws of physics are not violated, a PINN does not guarantee a physically valid solution. But it encourages the solution to be close to one.

To this end, we assume that our data is the product of a physical process that follows some dynamics which can be described by a PDE in the general form of
\begin{align}
  \mathcal{N}_{\vx}\left[u\right] &= 0, \qquad \quad\,\vx \in \Omega \subset \mathbb{R}^d \\
  u &= b(\vx_b), \quad \vx_b \in \partial \Omega, b : \partial \Omega \rightarrow \mathbb{R}\,,
\end{align}
with $u : \Omega \rightarrow \mathbb{R}$ being the solution operator, $ \mathcal{N}_{\vx} $ some potentially non-linear differential operator and $b$ a function describing boundary conditions.
For brevity, time-dependent components have not been noted explicitly in this depiction.
Given that this information is available, PINNs exploit it by substituting the solution $u$ with the prediction of the network $\fnetp$ and evaluating the differential operator using automatic differentiation to form a new physical loss term
\begin{equation}
  \ell_{p}(\vx;\theta) = \mathcal{N}_{\vx}\left[\fnetp\right]^2 \, .
\end{equation}
This is then added to the MSE loss function (\eqref{eq:loss}), resulting in a physics-informed loss
\begin{equation}
  \label{eq:loss_phys}
  \mathcal{L}_{p}(\theta) = \frac{1}{N} \sum_{i=1}^N \left( \vy_i - \fnet(\vx_i; \theta) \right)^2 + \ell_{p}(\vx;\theta) \, .
\end{equation} 
Both terms can be weighted by a factor that can be determined by, e.g., standard cross-validation.

\section{Reducing data demand}
\label{sec:sampling}
In this section we propose two ways to improve the standard collocation point sampling and loss function of PINNs. This enables the efficient use of PINNs in data-scarce simulation settings.
We assume that a fixed number of ``classical'' training data samples, which typically only consist of initial and boundary information, is available to compute the MSE loss.
We therefore target the number of collocation points as their use in the physics-based loss dominates the training time of the network: In each iteration of the PINN's training, the physical loss has to be evaluated at every collocation point.
And this, again, requires automatic differentiation. 
Thus, our goal is to reduce the number of collocation points while keeping the validation error at an acceptable level.

\subsection*{Penalizing Physical Loss Gradient}
Studying several benchmark problems with few collocation points, we noticed that when PINN predictions begin to fail, the optimizer typically finds a minimum where, at some point, the prediction starts to follow a trivial solution $u(\vx) = 0$ for the case of a homogeneous PDE.
From the point of view of the physics-informed penalty term, this makes perfectly sense since, depending on where the collocation points were drawn, the physical loss constraint is not violated. 
The reason is that the optimizer aims to satisfy the constraint at the collocation points where both the trivial and the correct solution are valid.
What happens in between these points however, is beyond the control of the PDE constraint.

However, as we will show, we have observed that the physical loss usually exhibits a spike or steep increase in the region where the prediction starts to follow a trivial solution.
To mitigate this, we propose to penalize the gradient of the physical loss. We propose to add a third penalty term to the loss function that penalizes the maximum gradient of the physical loss resulting in an additional term
\begin{equation}
  \label{eq:l_grad}
  \ell_{p_{grad}} = \max \nabla \ell_{p}(\vx;\theta) \, ,
\end{equation}
which is then added to the overall loss. We expect that penalizing spikes in the physical loss leads to predictions that are much less likely to adopt the trivial solution.

\subsection*{Latin Hypercube Sampling (LHS)}
The state-of-the-art of sampling collocation points for PINNs, as it was first proposed in \cite{pinn}, is to use Latin Hypercube sampling with a uniform distribution.
To give a rough idea, the domain $\Omega$ in which we want to sample is first partitioned into $N$ intervals of equal size in each dimension, where $N$ is the number of samples we want to draw.
In each of these intervals the final sample is now drawn from a uniform distribution. In higher-dimensional settings, each row or column of the resulting cartesian grid is only allowed to contain a single point.
Thus, this method can be understood as a compromise between pure random sampling and a grid-based distribution of points.

\subsection*{Regular Sampling}
While vanilla LHS provides a popular, flexible sampling strategy, its coverage of the domain can be less advantageous in scenarios with only few samples. We have therefore compared LHS with other sampling strategies. As we will show, a simple, equidistant, regular sampling can be superior to LHS for low-dimensional problems. This holds in particular if the optimization target is to reduce the number of points as much as possible.

\section{Experiments}
\label{sec:results}

To demonstrate the effect of the aforementioned variants, we use a PINN to predict the motion of a simple 1-D harmonic oscillator.
It is governed be the second-order ordinary differential equation (ODE)
\begin{equation}
  m\frac{\partial^2 u} {\partial t^2} = -ku \, ,
\end{equation}
with $m = 1.5, k=1.5$ chosen as mass and spring constant.

We start by using a fully-connected neural network of 8 layers with 20 artificial neurons each, with tanh activation functions.
The network is trained using the ADAM \cite{adam} optimizer with a default learning rate of $0.001$ and no additional learning rate scheduler.

As the governing equation is of second order in this example, we have to take an extra step for the initial condition.
To produce a unique solution, both an initial value and an initial tangent have to be defined.
For the first results, we set
\begin{align}
  u(0) &:= -2 \\
  \frac{\partial u}{\partial t}\left(0\right) &:= 0 \, .
\end{align}

Building a physical loss function for this experiment according to \eqref{eq:loss_phys}, leaves us with the overall loss for our example defined as
\begin{equation}
  \begin{split}
    \mathcal{L}_{p}(\theta) = \frac{1}{N} \sum_{i=1}^N \left[m\frac{\partial^2 \fnet(t_i; \theta)} {\partial t^2} + k\fnet(t_i; \theta)\right]^2 \\ + \left[\frac{\partial \fnet(0; \theta)} {\partial t}\right]^2 +  \left[-2 - \fnet(0; \theta) \right]^2 \, ,
  \end{split}
\end{equation}
where the sum requires satisfaction of the governing equation at $N$ collocation points and the last two terms require satisfaction of the second-order initial conditions.
It is worth noting, that in this experiment the initial conditions are the only ``true'' input-output data points that are known a priori.

\begin{figure}[ht]
  \begin{center}
     \centerline{\includegraphics[width=\columnwidth]{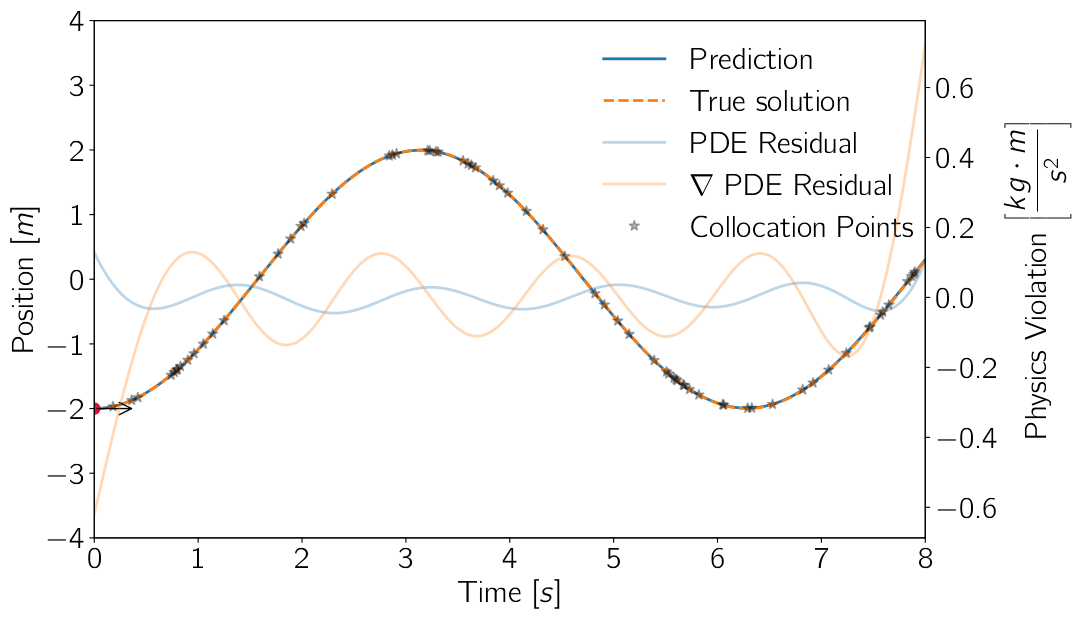}}
  \end{center}
  \caption{Solving the harmonic motion using a PINN. The only ``classical'' data point specifies the initial condition at $t=0$.}
  \label{fig:ham_osc}
\end{figure}

In Figure \ref{fig:ham_osc} we show the result of training a network using this approach with random sampling and penalizing the physical loss at $68$ collocation points.
After only $452$ epochs, the network is able to reproduce the true solution very accurately.

\begin{figure}[ht]
  \centering
     \centerline{\includegraphics[width=\columnwidth]{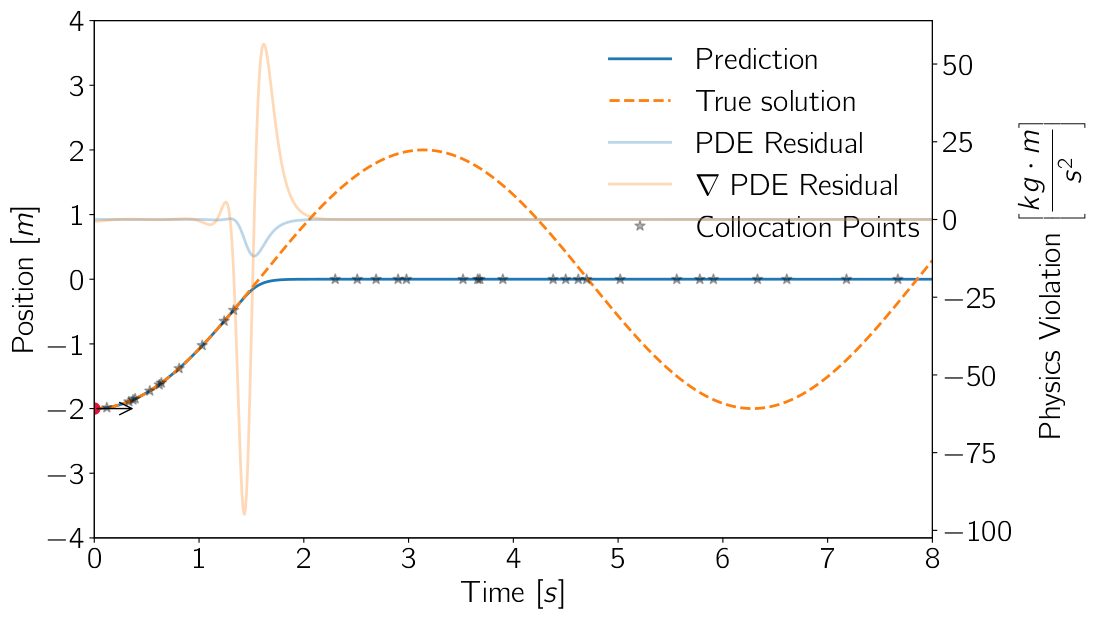}}
     \caption{The PINN fails with a reduced number of 32 collocation points. A peak in physical loss and its gradient is visible at the point of failure.}
     \label{fig:ham_osc_trivial}
\end{figure}

Remember that our goal is to reduce the number of collocation points as much as possible.
Doing so, we can clearly see that the prediction can become unstable and suddenly starts following the zero line as it satisfies the trivial solution, which can be observed in Figure \ref{fig:ham_osc_trivial}.
The switch to the constant trivial solution comes into effect where, due to random sampling, there is a larger gap between two subsequent collocation points.

Because of this, the ODE constraint does not have to be satisfied in this interval and the optimizer chooses a prediction that starts following the trivial solution for the rest of the prediction.
The result is that the overall physical loss gets reduced significantly and the chance of the optimizer leaving this local minimum becomes minimal.
This was observed by reducing the amount of collocation points to just $32$.
It is worth noting, however, that due to the random nature of optimizing NNs these results may vary. Even for such a low number of collocation points it can happen that in some rare cases the random method produces a sensible prediction.

We therefore add the additional penalty term as defined in \eqref{eq:l_grad} to the overall loss function and obtain the total loss

\begin{equation}
  \begin{aligned}
     \mathcal{L}_{p}(\theta) &= \frac{1}{N} \sum_{i=1}^N \left[ m\frac{\partial^2 \fnet(0; \theta)} {\partial t^2} + k\fnet(0; \theta)\right]^2 \\ &+ \max_{i=1,\ldots,N} \left[\left( m\frac{\partial^3 \fnet(t_i; \theta)} {\partial t^3} + k\frac{\partial\fnet(t_i; \theta)}{\partial t}\right)^2 \right] \\ &+ \left[\frac{\partial \fnet(0; \theta)} {\partial t}\right]^2 +  \left[-2 - \fnet(0; \theta) \right]^2 \, .
  \end{aligned}
\end{equation}

\begin{figure}[ht]
  \begin{center}
     \centerline{\includegraphics[width=\columnwidth]{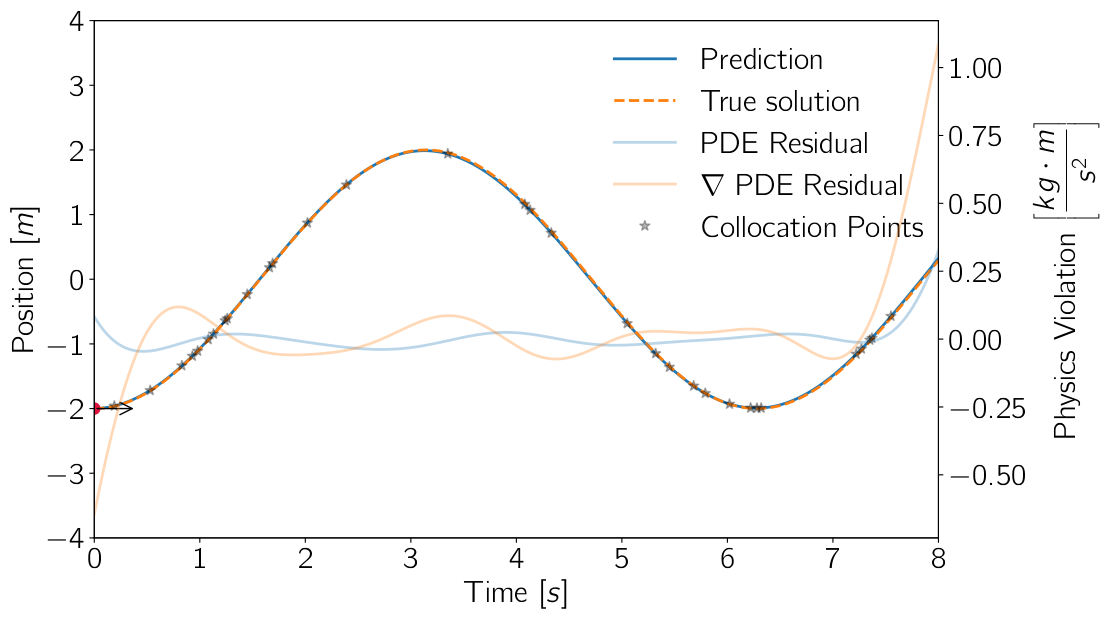}}
  \end{center}
  \caption{32 collocation points with physical loss gradient penalization.}
  \label{fig:ham_osc_pen}
\end{figure}

Training the model using this loss function with the same number of $32$ collocation points, a significant improvement can be observed in Figure \ref{fig:ham_osc_pen}.
The physical loss has clearly been smoothed out by penalizing the maximum of its gradient, resulting in the prediction not degenerating and jumping to the trivial solution as it did before.
Furthermore, it now fluctuates in a range that is an order of magnitude smaller than it was before.

\begin{figure}[ht]
  \begin{center}
     \centerline{\includegraphics[width=\columnwidth]{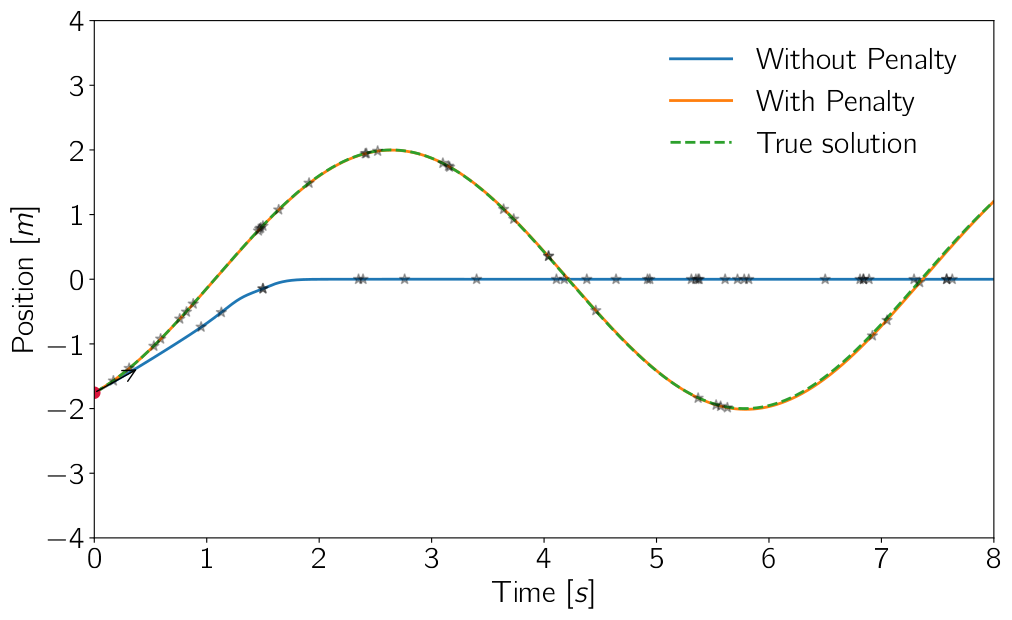}}
  \end{center}
  \caption{Prediction with and without physical loss gradient penalization for 32 collocation points and shifted initial conditions.}
  \label{fig:ham_osc_pen_regular_shifted}
\end{figure}

In order to demonstrate this behavior for a different setting we also trained a PINN network with the same amount of collocation points on a harmonic oscillator problem with slightly shifted initial conditions.
As shown in Figure \ref{fig:ham_osc_pen_regular_shifted}, the same behavior as before can be observed.
While the default approach fails and snaps to the trivial solution, the version with an additional penalty term is able to learn the correct solution.

\begin{figure}[ht]
  \begin{center}
     \centerline{\includegraphics[width=\columnwidth]{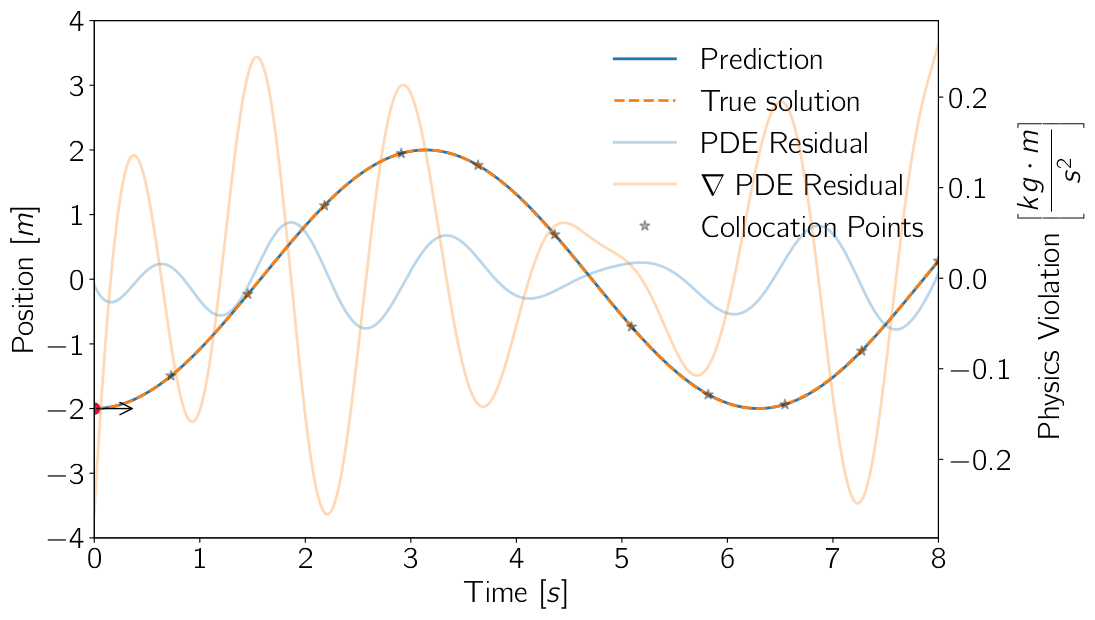}}
  \end{center}
  \caption{Prediction with just 12 collocation points using physical loss gradient penalization and regular grid sampling.}
  \label{fig:ham_osc_pen_regular}
\end{figure}

In an attempt to reduce the number of collocation points even further we now switch to regular grid sampling instead of random sampling.
As can be seen in Figure \ref{fig:ham_osc_pen_regular}, this has allowed the method to work with just 12 collocation points.
With such a low number of sampled points all other combinations of the proposed methods would fail in our experiments.

To consolidate the effects of regular grid sampling further, we did an extensive study by varying the number of collocation points $n_c$ from $10$ to $50$.
For each number of collocation points we trained the network $30$ times and calculated the resulting mean-squared error (MSE) to the analytical solution in $1000$ equidistantly placed points on the time domain.

\begin{figure}[ht]
  \begin{center}
     \centerline{\includegraphics[width=\columnwidth]{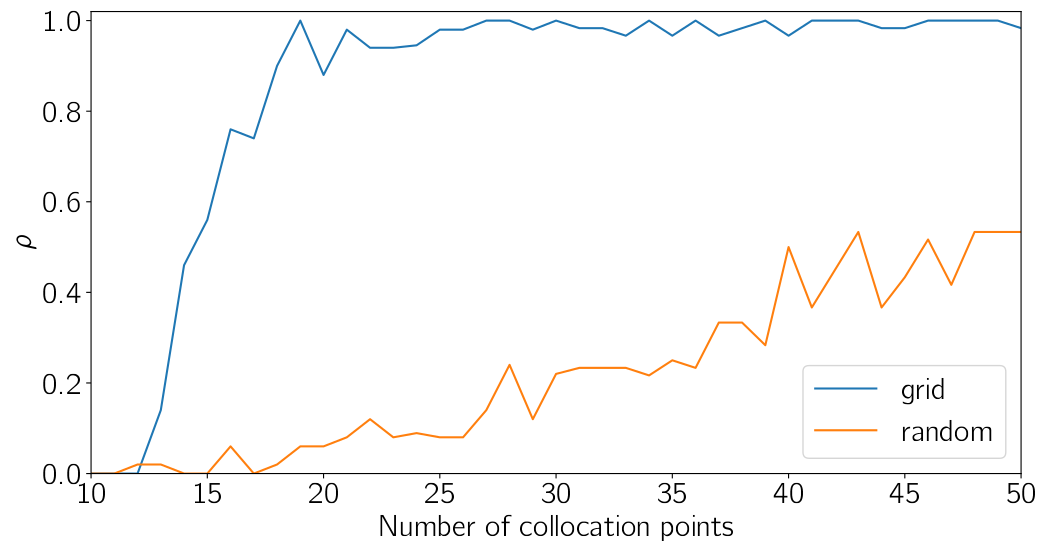}}
  \end{center}
  \caption{Ratio $\rho$ of successful runs out of $30$ training runs for both grid and random sampling plotted over number of collocation points.}
  \label{fig:study}
\end{figure}

Figure \ref{fig:study} shows the portion $\rho$ of runs that we deemed successful by having an MSE below $0.01$ for both regular grid and random sampling.
This may be interpreted as a measure of how likely it is that training with the given amount of collocation points will result in a reasonably good prediction.
From this figure, we clearly see that for regular grid sampling the ratio reaches $1$ much earlier than for the random sampling approach.
The difference is especially prominent for $n_c \approx 20$, where it is indicated that almost all of the regular grid sampling runs give pretty good predictions while most of the random sampling runs fail. This clearly demonstrates robustness of the prediction given constraints in their computational resources. 

\section{Conclusion}
\label{sec:conclusion}

Applying machine learning methods such as neural networks in a simulation environment usually requires considering data availability.
For physics-informed networks, this problem gets tackled already to some degree by requiring less traditional data and instead shifting the computational complexity towards evaluating a PDE constraint in the loss function of the network.
In this work we have investigated ways of improving physics-informed neural networks beyond their state-of-the-art. We have shown that the amount of collocation points at which the PDE constraint is evaluated can be significantly reduced.
We have proposed two ways of improving the data demand while still retaining an acceptable amount of prediction accuracy.

First, an additional penalty term is proposed which makes sure that the physical loss does not exhibit any sudden spikes which we have identified to be an indicator for bad predictive performance.
In our test case, this has resulted in the ability to reduce the number of needed collocation points by $\approx 50\%$.

Second, we suggest taking a simpler approach to sampling the collocation points at which the physical loss is constrained during training in data-scarce regimes.
Instead of using a random approach such as Latin Hypercube sampling we have examined placing the points on an equidistant grid.
In combination with the new, additional penalty term, this has made the difference between not being able to train a good prediction with just $12$ collocation points at all and getting a reasonably good result.

We have already observed that the observations studied here at the example of the one-dimensional, time-dependent harmonic oscillator also hold for more complex PDE problems such as the higher-dimensional wave equation or the Burgers equation. 
In future work, however, we will continue our studies in which we will test the proposed improvements on more complex problems and validate our findings.
We suspect the regular grid sampling to perform poorly in higher-dimensional settings. As a result we plan to experiment with different deterministic and pseudo-random sampling strategies such as Sparse Grid sampling or Sobol sequences.

\def\UrlBreaks{\do\/\do-}
\bibliography{bib}
\bibliographystyle{icml2021}





\end{document}